\documentclass[10pt,twocolumn,letterpaper]{article}
\usepackage{times}
\usepackage{epsfig}
\usepackage{graphicx}
\usepackage{amsmath}
\usepackage{amssymb}
\usepackage{authblk}
\usepackage{enumitem}

\usepackage[pagebackref=true,breaklinks=true,letterpaper=true,colorlinks,bookmarks=false]{hyperref}

\usepackage[utf8]{inputenc} 
\usepackage[T1]{fontenc}    
\usepackage{hyperref}       
\usepackage{url}            
\usepackage{booktabs}       
\usepackage{amsfonts,amsmath}       
\usepackage{nicefrac}       
\usepackage{microtype}      
\usepackage[algoruled
,boxed,lined]{algorithm2e}
\usepackage{graphicx}
\usepackage[export]{adjustbox}
\usepackage{xcolor}
\usepackage{colortbl}
\usepackage{array,colortbl,ctable}
\usepackage{subfig}
\newcommand{\suttl}[0]{MimicGAN}
\usepackage{siunitx}
\usepackage{color}
\usepackage{multirow}
\usepackage{listings}
\usepackage{color}
\usepackage{textcomp}
\usepackage{titling}

\usepackage[toc,page]{appendix}

\definecolor{listinggray}{gray}{1.0}
\definecolor{lbcolor}{rgb}{1.0,1.0,1.0}
\title{MimicGAN: Corruption-Mimicking for Blind Image Recovery \& Adversarial Defense}

\begin{document}
\setlength{\abovedisplayskip}{1pt}
\setlength{\belowdisplayskip}{2pt}

 \author{Rushil Anirudh, Jayaraman J. Thiagarajan, Bhavya Kailkhura, Timo Bremer \\ \{anirudh1, jjayaram, kailkhura1, bremer5\}@llnl.gov}
 \affil[]{Lawrence Livermore National Laboratory.}
			\date{}

\makeatletter
		\let\@oldmaketitle\@maketitle
		\renewcommand{\@maketitle}{\@oldmaketitle
			\vspace{-1cm} \centering	\includegraphics[trim={0cm 0cm 35cm 30cm},clip,width=0.8\linewidth]{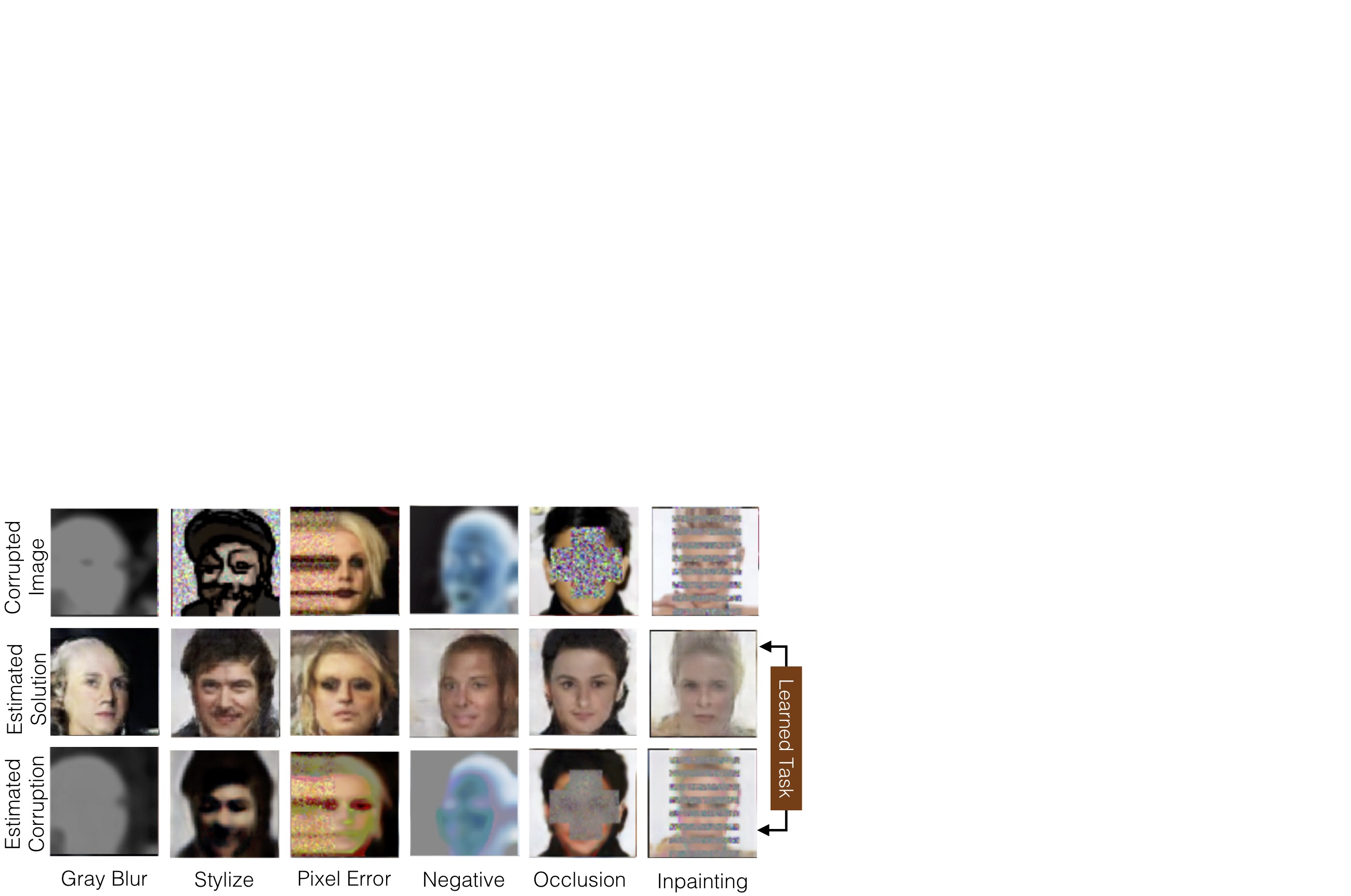}
			 \captionof{figure}{\suttl~is a fully unsupervised technique that can recover images from unknown corruptions by making successive estimates on the corruption as well as the solution. It can solve challenging inverse tasks without task-specific training (or data), and also outperform recent GAN-based defenses against several types of adversarial attacks.}
			 \label{fig:teaser}\bigskip}
		\makeatother
		\maketitle


\begin{abstract}
\emph{Solving inverse problems continues to be a central challenge in computer vision. Existing techniques either explicitly construct an inverse mapping using prior knowledge about the corruption, or learn the inverse directly using a large collection of examples. However, in practice, the nature of corruption may be unknown, and thus it is challenging to regularize the problem of inferring a plausible solution. On the other hand, collecting task-specific training data is tedious for known corruptions and impossible for unknown ones. We present MimicGAN, an unsupervised technique to solve general inverse problems based on image priors in the form of generative adversarial networks (GANs). Using a GAN prior, we show that one can reliably recover solutions to underdetermined inverse problems through a surrogate network that learns to mimic the corruption at test time. Our system successively estimates the corruption and the clean image without the need for supervisory training, while outperforming existing baselines in blind image recovery. We also demonstrate that MimicGAN improves upon recent GAN-based defenses against adversarial attacks and represents one of the strongest test-time defenses available today.}

\let\thefootnote\relax \footnotetext{This work was performed under the auspices of the U.S. Department of Energy by Lawrence Livermore National Laboratory under Contract DE-AC52-07NA27344.}
\end{abstract}

\section{Introduction}
Recovering good approximations of an image given its corrupted version is a widely studied problem in computer vision. It falls into a broader class of under-determined inverse problems, where a \emph{good} solution requires us to make informed guesses to compensate for missing information. Formally, this refers to problems that take the form $Y = f(X) + \eta$, wherein the goal is to recover a clean signal $X$ from its noisy ($\eta$ is a noise source), incomplete observation $Y$~\cite{ribes2008linear}. In computer vision, the mapping $f$ encompasses a wide range of deformation or corruption functions~\cite{xie2012image,pathak2016context,yeh2017semantic,chang2017one}. Existing solutions can be broadly divided into two categories: (a) \textit{Explicit} solutions which assume knowledge of the corruption model, $f$; and (b) \textit{Implicit} solutions that learn to approximate the inverse, $f^{\dagger}$, using a large set of $(X,Y)$ pairs. The former class of approaches are effective in practice, particularly when suitable priors on the signal space are available (e.g. $\ell_1$ or total variation). In cases where the priors are not known, recent approaches have adopted Generative Adversarial Networks (GANs)~\cite{GANGoodfellow} to regularize the inversion process with a \textit{GAN prior} instead~\cite{yeh2017semantic,pathak2016context,chang2017one}. While, explicit methods, both traditional and GAN based, typically do not require large databases of labeled examples, they do leverage information about the known corruption model which may not be available, i.e.\ in case of an unknown sensor or uncertain physics process.

\noindent In contrast, the implicit approaches rely entirely on data-driven inferencing (e.g. deep learning); examples include reconstruction from compressed measurements~\cite{chang2017one,Reconnet}, super-resolution~\cite{ledig2016photo}, or CT image recovery from incomplete views~\cite{anirudh2017lose} etc. While these implicit approaches eliminate the need to characterize the hypothesis space of corruptions, or to define appropriate signal priors, the resulting solutions are highly specific to the trained task, and their performance deteriorates when sufficient number of training observations are not available. Both explicit and implicit methods fail to deal with \emph{unknown} corruptions, a critical requirement when image classifiers are increasingly deployed in the \textit{wild}, where there  is no \textit{a priori} knowledge about the image distortions that can potentially make the classifier ineffective. Here, we seek a more realistic scenario where we neither have any knowledge about the corruption $f$, nor have access to training examples. In particular, we consider the important problems of blind image recovery and defending against adversarial attacks.

\noindent We present \suttl, an unsupervised approach for solving inverse problems, which utilizes a GAN prior while also learning the corruption on the fly. There are two main components in \suttl: (a) a pre-trained GAN that can act as a prior for the signal space; and (b) a \textit{surrogate} neural network that attempts to estimate $f$ given few observations, and the best guess from a GAN prior.  Note that, the GAN is trained once offline on clean images, while the surrogate only uses a small set of corrupted observations that are made available at test time. Therefore, we never need training pairs $(X,Y)$ which is a significant advantage, and results in a very general, task-agnostic approach. In contrast to implicit approaches that estimate the inverse mapping $f^{\dagger}$, we attempt to generate the corrupted observations by iteratively obtaining an approximation of the clean images from the GAN prior, and an estimate of the corruption from the surrogate, $Y \approx \hat{f} \hat{X}$. The inclusion of a corruption-mimicking surrogate into the inversion process acts as a regularizer when estimating the clean image using the GAN. Surprisingly, this alternating optimization is highly effective even for complex corruption functions, which existing GAN prior-based methods fail to invert. Figure~\ref{fig:teaser} illustrates the effectiveness of \suttl~in a large class of blind image recovery problems. Using only a small number of observations (as low as $25$) at test time, \suttl~is able to determine highly plausible solutions, while reasonably mimicking the actual corruption process.
\vspace{5pt}

\noindent Another important application of \suttl's ability to correct for unknown corruptions is in generating defenses against adversarial attacks~\cite{GoodfellowSS14}, i.e.\ quasi-imperceptible perturbations designed to fool a classifier. We pose adversarial defense as an image recovery problem, where we are given images that have been altered by very specific perturbations. We report that for several popular attacks, the \suttl~defense achieves state-of-the-art performance, when compared to existing GAN-based defenses~\cite{defenseGAN,ilyas2017robust}.
\vspace{5pt}

\noindent \textbf{Impact:} To the best of our knowledge, \suttl~is the first system capable of effectively recovering images from such a diverse class of corruptions, without the need to collect a large number of observations. While \suttl~eliminates the need to explicitly define suitable signal priors, it also does not require knowledge about the corruption, taking us a step closer towards a \emph{universal} solution for inverse problems. This can eventually lead to the design of image recovery systems that can be deployed as a pre-processing step for achieving robust classification in the \textit{wild}.

\vspace{5pt}

\noindent
Our contributions can be summarized as follows:
\begin{itemize}[leftmargin=*,itemsep=-1.mm]
  \item[1.] We propose \suttl, the first fully unsupervised approach to solve challenging inverse problems, \emph{without prior knowledge of the corruptions}.
  \item[2.] We introduce a corruption-mimicking surrogate to the inversion, improving the performance of existing GAN based methods without additional supervision or data.
  \item[3.] We demonstrate \suttl's effectiveness on six challenging blind image recovery tasks using the CelebA Faces dataset~\cite{liu2015faceattributes} with competitive results compared to fully supervised techniques and significantly better results than unsupervised baselines.
  \item[4.]  On the Fashion-MNIST~\cite{fashion-mnist} dataset we show that \suttl~significantly improves the robustness of pre-trained classifiers to extreme corruptions.
  \item[5.] Finally, we present the \suttl~defense against adversarial attacks on the Fashion-MNIST dataset and show that it is the best test-time defense, against six of the strongest attacks available today.
\end{itemize}


\section{Related Work}
\begin{figure*}[!htb]
	\centering
	\subfloat[\suttl~ solves image recovery by alternating between the corruption and the solution.]{
		\includegraphics[trim={7cm 8cm 5cm 7cm},clip,width=0.65\linewidth,valign=c]{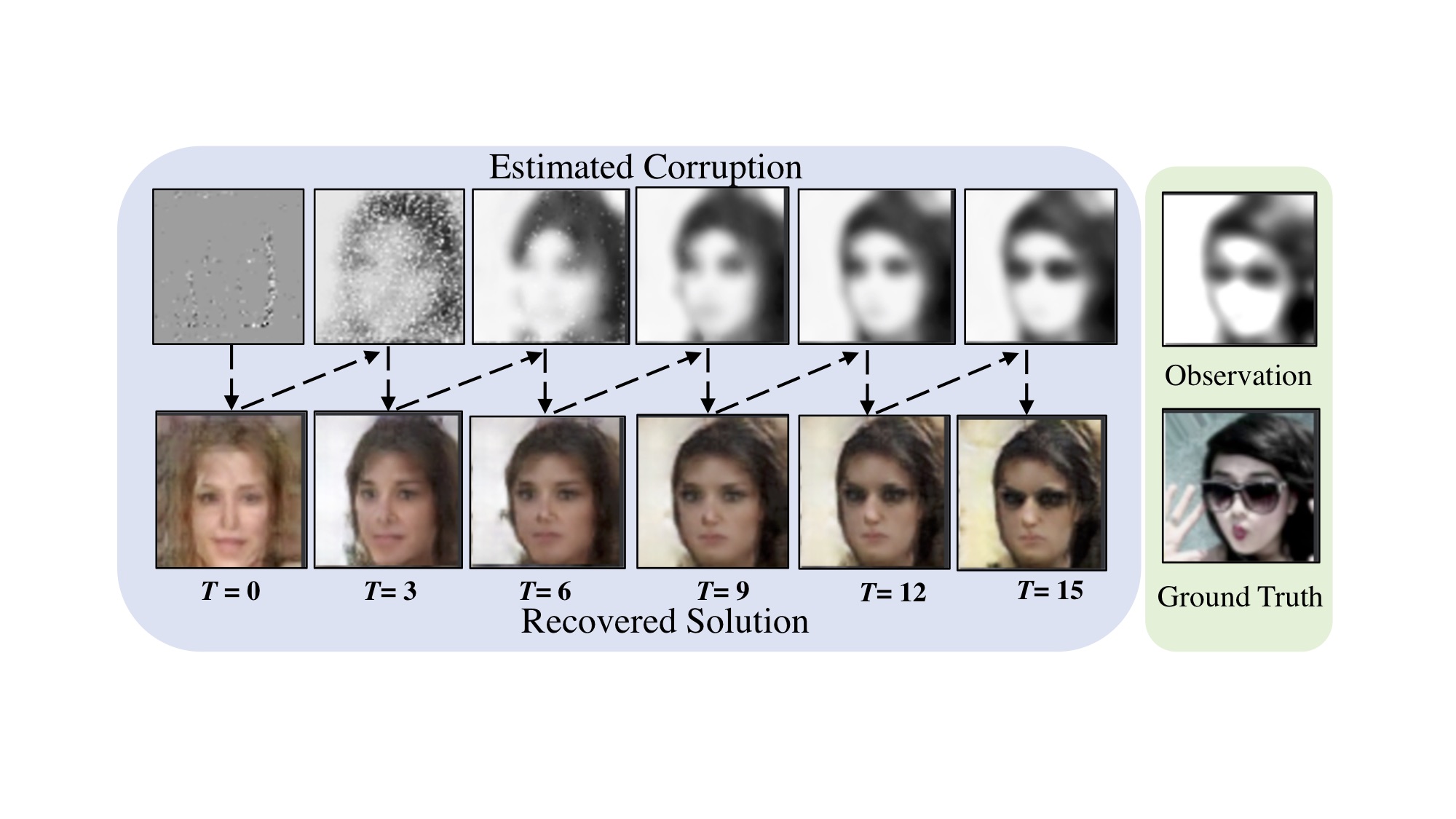}
		\label{fig:alternating}}
	\subfloat[Quality of estimated solution vs. number of alternating optimization steps.]{
		\includegraphics[trim={0 1cm 0cm 2cm},clip,width=0.28\linewidth,valign=c]{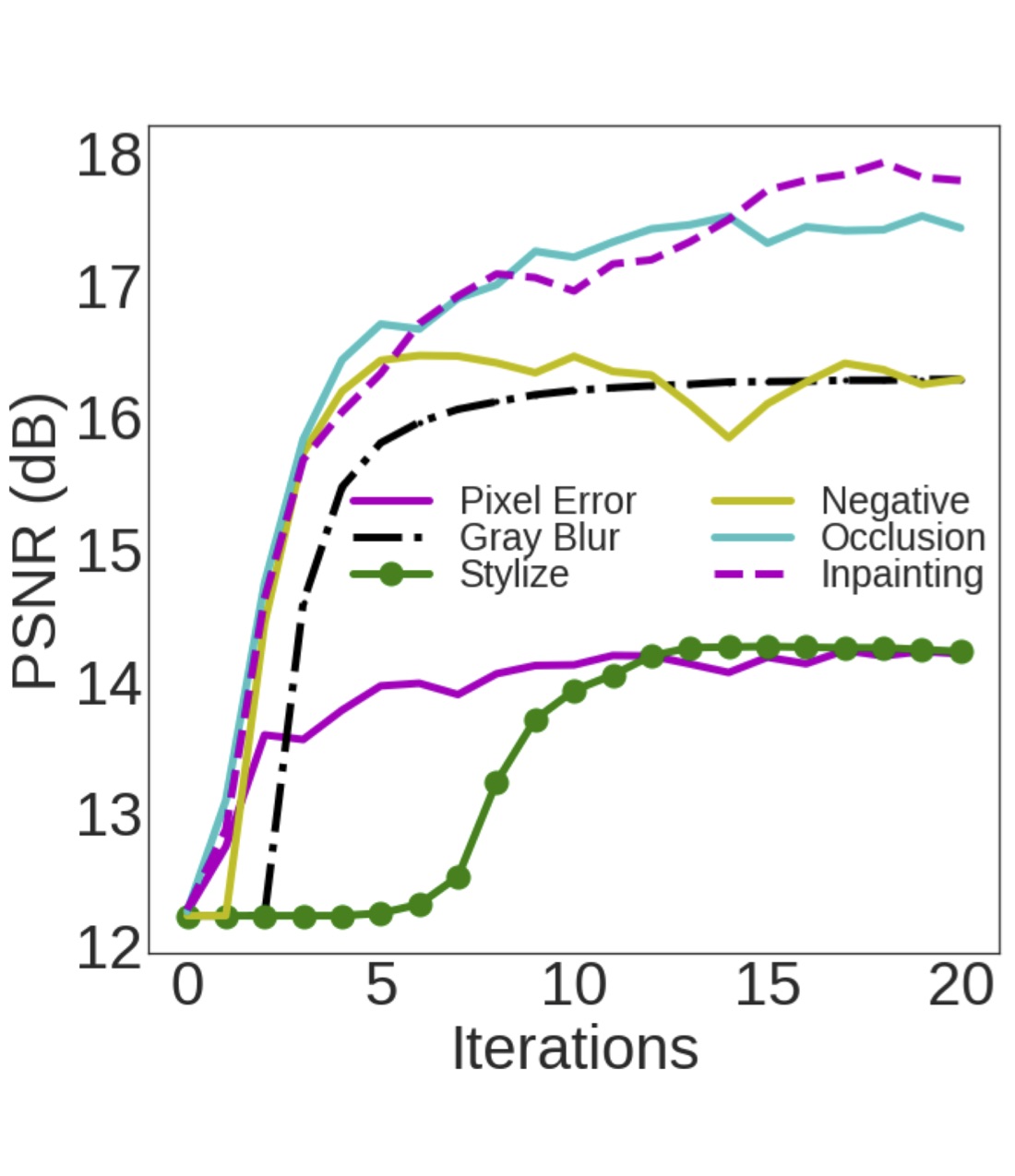}
		\label{fig:psnr}}
	\caption{A demonstration of the proposed approach. We iteratively estimate the corruption and guess images from the GAN. As the surrogate captures finer details (such as eye glasses), \suttl~ correspondingly adds them to the estimated solution.}
	\label{fig:evolve}
	\vspace{-10pt}
\end{figure*}
\noindent \textbf{GANs for Inverse Problems:} With their ability to effectively model high-dimensional distributions, GANs~\cite{GANGoodfellow} have enabled the use of unsupervised methods in inverse problems. Yeh \textit{et al.}~\cite{yeh2017semantic} first introduced the projected gradient descent (PGD) method with GANs for filling up arbitrary holes in an image, in a semantically meaningful manner. Subsequently, several efforts have pursued solving inverse problems using PGD, for example compressive recovery~\cite{bora2017compressed, shah2018solving}, and deblurring~\cite{AsimDeblurGAN}. Asim \textit{et al.}~\cite{AsimDeblurGAN} proposed an alternative approach for image deblurring, wherein they used two separate GANs -- one for the blur kernel, and another for the images. {\suttl} differs from this approach in that we make no assumptions on the types of corruptions that can be recovered, and hence requires only a single GAN for modeling the image manifold. The corruption-mimicking proposed in this paper is loosely related to the cyclical consistency used in the CycleGAN architecture~ \cite{isola2017image}; however with a crucial difference that~\cite{isola2017image} infers the forward and inverse mappings between the two distributions together, thus requiring large datasets in both domains. In contrast, our approach focuses on image recovery scenarios, where very few examples (for e.g. $\sim 25$) are available.

\noindent AmbientGAN \cite{bora2018ambientgan}, is another related technique that allows one to obtain a GAN in the original space given its lossy measurements, which can be useful to obtain a GAN prior even when clean images are not available. However, the authors in~\cite{bora2018ambientgan} consider only inverse problems where the \emph{form} of corruption function is known \textit{a priori} (e.g. random binary mask). Whereas, {\suttl} uses a surrogate network to emulate the corruption process, and can be applied to observations from any arbitrary corruption function. Finally, our work improves on the notion of ``GAN priors" \cite{yeh2017semantic,shah2018solving, AsimDeblurGAN, bora2017compressed} -- the idea that optimizing in the latent space of a pre-trained GAN provides a powerful prior to solving several traditionally hard problems. As GANs become better, we expect new capabilities in solving challenging inverse problems to emerge.

\noindent \textbf{GANs for Adversarial Defense:} In addition to its effectiveness for image recovery, the GAN-prior has also become a reasonably successful way to defend against adversarial attacks -- which are small perturbations in the pixel space that are designed to cause a particular classifier to fail dramatically. For example, Defense-GAN \cite{defenseGAN}, uses an approach called \textit{clean and classify}, where the observed adversarial example is projected on to the image manifold, with the hope of eliminating the adversarial perturbation in the process. A very related idea is explored in \cite{ilyas2017robust}, referred as \textit{Invert and Classify} (INC), which also relies on PGD as in the case of Defense-GAN. Note that, both Defense-GAN and {\suttl}are applicable to white-box as well as black-box attacks, since they do not need access to the classifier model in order to clean the data. Another related approach proposes to use the discriminator in addition to the generator to detect adversarial examples \cite{santhanam2018defending}, and we show in our experiments that, in comparison, the \suttl~defense is significantly more robust. Interesting, all the three aforementioned approaches become a special case of {\suttl}, when the surrogate network in our approach is assumed to be identity. It should be noted that most existing GAN based defenses are effective only when the attack is designed purely on the classifier. However, recent evidence shows that these defenses can be broken when the attack is designed on the GAN and the classifier together \cite{athalye2018obfuscated}. We observe that, the \suttl~defense is also vulnerable to such a GAN-based attack, but to a lower degree than plain GAN-based defenses, and this can be attributed to inclusion of the corruption-mimicking process.


\section{Proposed Approach}
\label{sec:proposed}
In this section, we present details of \suttl, an entirely unsupervised approach that uses a surrogate to compensate for the lack of knowledge about $f$, and a pre-trained generative model (GAN) to determine plausible solutions on the image manifold.
\vspace{5pt}

\noindent \textbf{Problem Formulation:} Let us denote a set of $N$  observed images as $\mathbf{Y}^{obs}\in \mathbb{R}^{N \times d}$ where each column $Y_j^{obs} \in \mathbb{R}^{d}$ denotes an independent observation. We assume that each $Y_j^{obs}$ is a corrupted version of the clean image $X_j \in \mathcal{X} \subset \mathbb{R}^{M}$. Here $\mathcal{X}$ denotes the image space, and without loss of generality, we assume $d \leq M$. The observations are obtained through an unknown corruption $f$, i.e., $Y_j^{obs} = f\left(X_j \right)+\eta$. Here $\eta$ represents noise in the measurement process. The goal is to recover an estimate of the clean image $\hat{X}_j$ from its observation $Y_j^{obs}$ without any knowledge of $f$, and very few observations.

\noindent When the corruption $f$ is known the recovery problem can be formulated as:
\begin{equation}
\label{eq:classic}
\underset{\{X_j \in \mathbb{R}^{M}\}_{j=1}^N}{\arg\min} \;
 \sum_{j = 1}^N \mathcal{L}\left(Y_j^{obs}, f(X_j)\right) + \lambda \mathcal{R} (X_j),
\end{equation}
where $\mathcal{L}$ denotes the loss function, $\mathcal{R}$ is an analytical regularizer and $\lambda$ is a penalty for the regularization term. The hope of solving a regularized optimization problem is that regularization $\mathcal{R}$ will lead to a plausible solution in $\mathcal{X}$. Recently, this regularization has been effectively realized using a GAN, referred to as the \textit{GAN prior} \cite{AsimDeblurGAN,bora2017compressed, shah2018solving,yeh2017semantic}, which restricts the solution to lie on a known image manifold. Formally, this can be expressed as $\mathcal{G}: z \mapsto X$, where $\mathcal{G}$ denotes the generator from a pre-trained GAN and $z \in \mathbb{R}^K$ is the latent noise vector.

\subsection{Corruption-Mimicking as a Regularizer}
\suttl~is a test-time only system that can solve inverse problems without any supervision. It uses a neural network to mimic the unknown corruption process $f$, with a plausible estimate of the clean image in $\mathcal{X}$ (parameterized by $\mathcal{G}$). Denoting the estimated corruption from the surrogate by $\hat{f}$, we hypothesize that $\hat{f}$ can effectively regularize the search on the image manifold and help recover a better clean image. Similarly, an estimate of the clean image $\hat{X}_j$ from $\mathcal{G}$ can be utilized to update the network parameters of the surrogate. This alternating optimization progresses by incrementally refining the surrogate, while ensuring that there always exists a valid solution $\hat{X}_j \in \mathcal{X}$, such that $\hat{f}(\hat{X}_j) \approx Y_j^{obs}$. The networks in the GAN are kept frozen in the entire optimization loop. Figure \ref{fig:alternating} demonstrates this process for an example case, and Figure \ref{fig:psnr} shows how the estimated solution improves, in terms of PSNR, with each iteration of the alternating optimization loop.

Note that, in contrast to other data-driven inversion methods, we approximate the forward process $f$ instead of the inverse $f^{\dagger}$. This formulation is applicable regardless of the form of $f$, and can be generalized to a large class of corruptions on the fly. Mathematically, we reformulate \eqref{eq:classic} as follows:

\begin{equation}
\label{eq:obj}
\hat{f},\{\hat{z}_j\}_{j=1}^N = \underset{{f},\{z_j\in \mathbb{R}^{N \times K}\}_{j=1}^N}{\arg\min} \;
\sum_{j = 1}^N \mathcal{L}\left(Y_j^{obs}, {f}(\mathcal{G} (z_j))\right),
\end{equation}where we have $\hat{X}_j = \mathcal{G}(\hat{z}_j)$. Since both $\mathcal{G}$ and $f$ are differentiable, we can evaluate the gradients of the objective in \eqref{eq:obj}, using backpropagation and use existing gradient based optimizers. In addition to computing gradients with respect to $z_j$, we also perform clipping in order to restrict it within the desired range (e.g., $[-1,1]$) resulting in a projected gradient descent (PGD) optimization. Solving this alternating optimization problem produces high quality estimates for both surrogate, $\hat{f}$ and the latent vectors for the recovered images, $\hat{X}_j = \mathcal{G}(\hat{z}_j)~\forall j$.

\vspace{5pt}

\noindent \textbf{Architecture:} The number of observations $N$ required by \suttl~depends on the corruption $f$, but for all the corruptions considered here, $N$ can be as low as $25$ samples. We implement the surrogate using a shallow network with 3 layers and 16 convolutional filters of size $5 \times 5$ in each layer. We use ReLU activations and a final masking operation that is also learned (see supplementary for details). Given the highly underdetermined nature of these inverse problems, in many cases the surrogate does not emulate the corruption exactly. Nevertheless, including the surrogate sufficiently regularizes the inversion. The balance between updating the surrogate and re-estimating the clean image using PGD is crucial for the algorithm to converge. Since the surrogate is updated using a  different set of images in each iteration (due to updates in $\hat{z}_j$), it does not overfit easily. However, we note that, overfitting can occur when: (a) the surrogate is very deep,  or (b) the balance between the two optimization loops is poorly chosen. In such cases, the surrogate causes \eqref{eq:obj} to converge prematurely, without providing useful gradients to update the GAN loop and leading to poor recovery.

\vspace{5pt}

\noindent \textbf{Losses:} Next we describe the construction of loss function $\mathcal{L}$ in (\ref{eq:obj}), which consists of three different losses:
\begin{itemize}

  \setlength\itemsep{-0.3em}
\item[(a)] \textbf{Corruption mimicking loss:} Measures the discrepancy between the observation $Y_j^{obs}$ and surrogate's prediction, $\mathcal{L}_{obs} =  \sum_{j = 1}^N \left\|Y_j^{obs}-{f}(\mathcal{G} ({z}_j))\right\|_1$, where $\|.\|_1$ is the $\ell_1$ norm.
\item[(b)] \textbf{Adversarial loss:} Using the discriminator $\mathcal{D}$ and the generator $\mathcal{G}$ from the pre-trained GAN, we measure: $\mathcal{L}_{adv} = \sum_{j=1}^N \log(1-\mathcal{D}(\mathcal{G}({z}_j))$. Note, this is the same as the generator loss used for training a GAN.
\item[(c)]\textbf{Identity loss}: In many cases such as inpainting, only a part of the original image is corrupted, and this loss exploits that information to guide $\hat{f}$ to converge closer to an identity transformation: $\mathcal{L}_{\mathcal{I}} = \sum_{j = 1}^N \left\{ \|Y_j^{obs}-\mathcal{G} (z_j)\|_1+\|{f}(\mathcal{G} (z_j))-\mathcal{G} (z_j)\|\right \}$. Naturally, this does not always help and can be avoided in cases where no part of the observed image exactly reproduces the clean image.
\end{itemize}Our overall loss is hence defined as:
\begin{equation}
  \label{eq:losses}
  \mathcal{L} = \mathcal{L}_{obs} + \lambda_{adv}\mathcal{L}_{adv} +  \lambda_{\mathcal{I}} \mathcal{L}_{\mathcal{I}} ,
\end{equation}where $\lambda_{\mathcal{I}}, \lambda_{adv}$ are hyperparameters and can vary depending on the corruption type. Note that, the projected gradient descent (PGD) technique~\cite{yeh2017semantic,defenseGAN} can be viewed as a special case of our approach, when ${f} = \mathbb{I}$, where $\mathbb{I}$ is identity, and $\lambda_{\mathcal{I}} = 0$.
\vspace{5pt}

\noindent \textbf{Initialization:} \suttl~depends on an initial seed to begin the alternating optimization, and we observed variations in convergence behavior due to the choice of a poor seed. In order to avoid this, we initialize the estimate of the clean images by computing an average sample on the image manifold, by averaging $1000$ samples drawn from the random uniform distribution. Note that, we initialize the estimate for all observations with the same mean image. We observe that this not only speeds up convergence, but is also stable across several random seeds.
\vspace{5pt}

\noindent \textbf{Algorithm:} The procedure to perform the alternating optimization is shown in Algorithm \ref{mainAlg}. We run the inner loops for updating the surrogate and $z_j$ for $T_1$ and $T_2$ iterations respectively. The projection operation denoted by $\mathcal{P}$ is the clipping operation, where we restrict the $z_j$'s to lie within $[-1,1]$. We use the RMSProp Optimizer to perform the gradient descent step in each case, with learning rates of $\gamma_s, \gamma_g$ for the surrogate update and GAN sample update respectively. Note that, since our approach requires only the observations $\mathbf{Y}^{obs}$ to perform inversion, it lends itself to a task-agnostic inference wherein the user does not need to specify the type of corruption or collect a large dataset \emph{a priori}.

\begin{algorithm}[!htb]
\SetKwFunction{GAN}{GAN}
\SetKwFunction{random}{random}
\SetKwInOut{Input}{Input}
\SetKwInOut{Output}{Output}
\SetKwInOut{Initialize}{Initialize}
\Input{ Observations $\mathbf{Y}^{obs} \in \mathbb{R}^{N\times d}$, Pre-trained generator $\mathcal{G}$ and discriminator $\mathcal{D}$.}
\Output{ Recovered Images $\hat{\mathbf{X}} \in \mathbb{R}^{N \times M}$, Surrogate $\hat{f}$}
\Initialize{ For all $j, \hat{z}_j$ is set as average of $1000$ realizations drawn from $~~\mathcal{U}(-1,1)$\tcp{see text}}
\Initialize{Random initialization of surrogate parameters, $\hat{\mathbf{\Theta}}$}

\BlankLine

\For{$t\leftarrow 1$ \KwTo $T$}{
    \BlankLine

  \For{$t_1\leftarrow 1$ \KwTo $T_1$  \tcp{update surrogate}}{
    $Y_j^{est} \leftarrow \hat{f}\left(\mathcal{G}(\hat{z}_j); \hat{\mathbf{\Theta}}\right),~\forall j$;

	Compute loss $\mathcal{L}$ using \eqref{eq:losses};

    $\hat{\mathbf{\Theta}} \leftarrow \hat{\mathbf{\Theta}} - \gamma_s~\nabla_\mathbf{\Theta}(\mathcal{L}) $;
    }

    \For{$t_2\leftarrow 1$ \KwTo $T_2$ \tcp{perform inversion}}{
      $Y_j^{est} \leftarrow \hat{f}\left(\mathcal{G}(\hat{z}_j);\hat{\mathbf{\Theta}}\right),~ \forall j$;

	Compute loss $\mathcal{L}$ using \eqref{eq:losses};

      $\hat{z}_j \leftarrow \hat{z}_j - \gamma_g~\nabla_z(\mathcal{L}),~ \forall j$;

      $\hat{z}_j \leftarrow \mathcal{P}\left(\hat{z}_j\right)~ \forall j$;
      }
      \BlankLine

  }
  return $\hat{f}$, $\hat{{X}}_j = \mathcal{G}(\hat{z}_j),~\forall j$.
\caption\suttl\label{mainAlg}
\end{algorithm}

\begin{figure*}[!h]
\centering
\includegraphics[trim={0cm 6cm 0cm 0cm},clip,height=0.9\textheight]{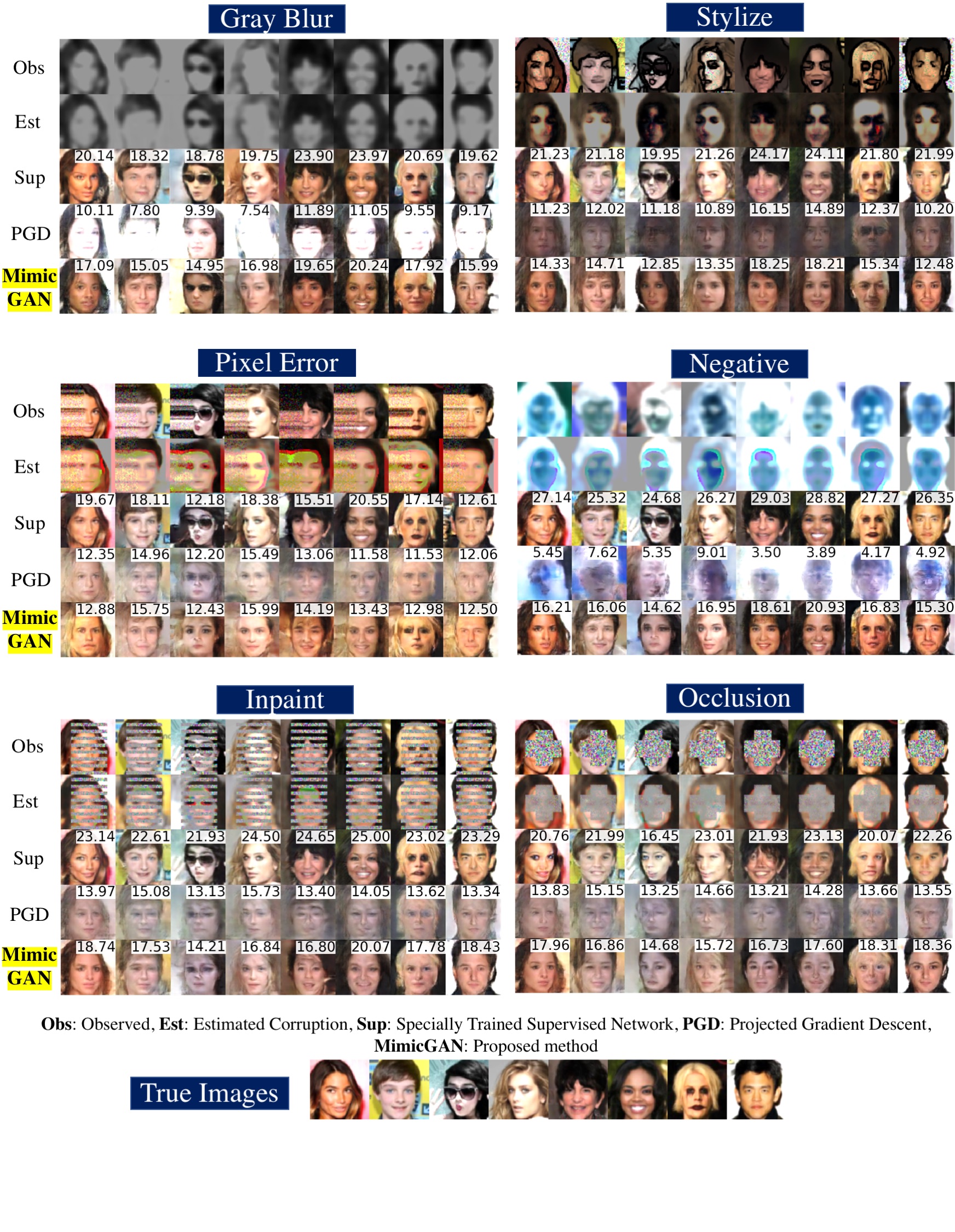}
\caption{\small{Image Recovery on the CelebA dataset with \suttl. Six tasks are shown here, in each case the rows represent the following: Observed, Estimated Corruption, Supervised Solution, PGD, Proposed. With no additional training or data requirements, we show an improvement of around $\sim 3-4$dB improvement over PGD.}}
\label{fig:celeba}
\end{figure*}
\section{Experiments}
In this section we present evaluations for \suttl~in blind image recovery and adversarial defense, and demonstrate its superiority over existing approaches. All our experiments are implemented using TensorFlow \cite{abadi2016tensorflow}.

\subsection{Blind Image Recovery}
Blind image recovery is the process of jointly estimating both the original image and the corruption, with no prior knowledge about the corruption process. Consequently, most existing solutions that are tuned for a specific corruption and cannot do blind image recovery. \suttl~does not make any such assumptions, and hence is broadly applicable to several types of corruptions. Here, we consider $6$ challenging image corruptions, which are described briefly next. Details on their implementations can be found in the supplementary material. We have chosen these representative tasks to demonstrate the ability of \suttl~in recovering highly plausible solutions to severely ill-conditioned problems; however we believe \suttl~can effectively recover a very large class of linear and non-linear image deformations.

The tasks considered here are:
\begin{itemize}
  \setlength\itemsep{-0.3em}
\item[1.] \emph{Gray Blur}: Extreme blur (scale = $25$), followed by converting the RGB image to grayscale.
\item[2.] \emph{Stylize}: OpenCV's stylization filtering, followed by adding Gaussian noise to top $10\%$ of the brightest image pixels ($\sigma = 0.5$).
\item[3.] \emph{Stuck Pixels}: Choose a column close to the middle, and replace all preceding columns with it. In addition, we add Gaussian noise to the pixels that have been replaced ($\sigma = 0.5$).
\item[4.] \emph{Negative}: Blur the image (scale = $11$), followed up inverting the RGB values.
\item[5.] \emph{Extreme Occlusion}: Cover the regions of face, eyes, nose and mouth with Gaussian noise ($\sigma = 0.5$).
\item[6.] \emph{Inpainting}: We overlay a pattern of lines (width = 4 pixels) on the image, with values drawn from a Gaussian distribution ($\sigma = 0.5$).
\end{itemize}
In all the scenarios, we assume no knowledge of the measurement process, and attempt to estimate the surrogate $\hat{f}$ and the clean signal jointly.

\noindent \textbf{Experimental Setup:}
All our image recovery experiments were carried out using the CelebA dataset~\cite{liu2015faceattributes}, which contains 202,599 images of size $64\times64$. We trained a DCGAN~\cite{radford2015unsupervised} $90\%$ of the images from CelebA as the GAN inside \suttl. We also rescale all the images to be in the range of $[-1,1]$. The validation set consisting of $10\%$ of the dataset is used in all the recovery experiments. Details on all the model architectures are available in the supplementary material. All recovery experiments for \suttl~were carried out with $N = 25$ observations. The hyperparameters for all the tasks are shown in table \ref{tab:hyperparam}. For the \emph{Extreme Occlusion} and \emph{Inpainting} tasks, we use TanH in the last layer instead of the ReLU activation.

\begin{table}[!htb]
\centering
{\small
	\begin{tabular}{p{2.3cm}p{1cm}p{1cm}p{1cm}p{1cm}}

		\hline
		\textbf{Task}  & $\lambda_{\mathcal{I}}$&$\lambda_{adv}$& $\gamma_s$& $\gamma_g$ \\ \hline
   Gray Blur   &0 & 0	 & $2\mathrm{e}{-4}$ & $1\mathrm{e}{-3}$    \\
	Stylize	 	& 0 & 0 & $8\mathrm{e}{-5}$ &  $4\mathrm{e}{-4}$\\
	Negative	& 0& $1\mathrm{e}{-4}$ & $1\mathrm{e}{-3}$ & $1\mathrm{e}{-3}$ \\
	Stuck Pixels & $1\mathrm{e}{-1}$& $1\mathrm{e}{-4}$ & $4\mathrm{e}{-3}$ &  $3\mathrm{e}{-3}$  \\
	Occlusion & $1\mathrm{e}{-1}$& $1\mathrm{e}{-4}$ & $4\mathrm{e}{-3}$ &  $3\mathrm{e}{-3}$  \\
	Inpainting & $1\mathrm{e}{-1}$& $1\mathrm{e}{-4}$ & $4\mathrm{e}{-3}$ &  $3\mathrm{e}{-3}$  \\\hline
\end{tabular}
\vspace{-5pt}
}
	\caption{\small{\textbf{Hyperparameters:} See section \ref{sec:proposed} for details on each parameter.$ \lambda_\mathcal{I}, \lambda_{adv}$ are regularization parameters (see \eqref{eq:losses}), and $\gamma_s, \gamma_g$ are the learning rates for the surrogate and finding samples on the GAN, respectively (See alg \ref{mainAlg}).}}
\label{tab:hyperparam}
\vspace{-10pt}
\end{table}

\noindent \textbf{Baselines:} We compare our technique with the following baselines: (a) \emph{Supervised:} We train a separate network for each recovery task, using paired corrupted-clean data. During training, we include an adversarial loss with $\lambda_{adv} = 1\mathrm{e}{-3}$ in order to produce more realistic solutions. Though supervised methods are known to be highly effective, their requirement for enough labeled examples can be unrealistic. Images recovered using our unsupervised approach are close to those obtained by the supervised counterparts; (b) \emph{PGD:} Projected Gradient Descent (PGD) used in \cite{AsimDeblurGAN,bora2017compressed, shah2018solving,yeh2017semantic} is a special case of our method when the $f$ is considered to be identity, and hence \suttl~is lower-bounded by the performance of PGD. While this baseline is surprisingly effective with a variety of corruptions, we find that it is easily susceptible to failure.
\vspace{5pt}

\noindent \textbf{Results:} Examples of the recovered images from all the tasks are shown in Figure \ref{fig:celeba}. Along with the reconstructions, we also show PSNR values for each case. It should be noted that PSNR is not a reliable performance metric and can be uncorrelated with the visual perception. However, we still significantly outperform PGD in terms of PSNR, while producing visually good quality reconstructions. While existing approaches rely predominantly on the GAN prior to perform recovery, we show that \suttl~shows significant performance gains.


\subsection{Robust Classification}
\label{sec:robust}
In several scenarios considered here, PSNR can be a poor performance metric as it does not capture the ability of \suttl~to recover important information or perceptual quality. Consequently, we quantify the improvement of our approach using a classification in the wild experiment. More specifically, we measure how well \suttl~is able to recover key attribute information which is useful for classification. This scenario is becoming increasingly relevant as classifiers are deployed in the wild, with little or no control over the test distribution. We perform these experiments on the Fashion-MNIST \cite{fashion-mnist} dataset, where we first train a GAN with a similar configuration as DCGAN (see supplementary for architectural details). Next, using the same train/validation split used for training the GAN, we train a convolutional neural network (CNN) classifier which achieves $91.50\%$ accuracy on a clean test set.

\begin{figure}[!htb]
	\centering

	\subfloat[A classifier is trained on the original/clean data and tested under different corruptions shown here]{\includegraphics[trim={1.0cm 2cm 1.5cm 2.5cm},clip,width=0.9\linewidth]{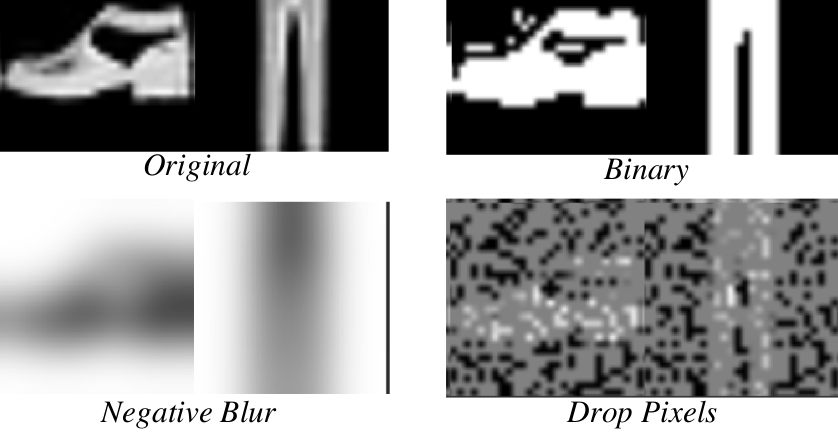}\label{fig:mnist_tasks}}\quad

	\subfloat[][Classification accuracy when test set is corrupted. Test performance on clean data is $91.50$.]{
	\small
	\begin{tabular}{p{2.2cm}p{1.1cm}p{0.5cm}p{1.1cm}}
		\hline
													& Observed 		& PGD						& \suttl\hspace{1pt}\emph{(Ours)} \\ \hline
\textbf{Binary}       		& 37.00			  & 25.50         & \textbf{50.40}    \\
\textbf{Dropped Pixels}	 	& 25.50 		  & 15.60         & \textbf{46.40}		\\
\textbf{Negative Blur}		& 7.20 				& 8.40 					& \textbf{45.80}  \\ \hline

	\end{tabular}
	\vspace{-10pt}
	\label{tab:mnist_robust}}

  \caption{\small{\textbf{Robust Classification:} As a pre-processing step, \suttl~can make classifiers more robust to arbitrary, unknown corruptions in the data. Here we report improvement in accuracy for three different, challenging scenarios.}}
  \label{tab:robust}
	\vspace{-5pt}
\end{figure}

\noindent Since \suttl~does not require any additional training or data, it can be used as an unsupervised pre-processing step to ``clean and predict'' labels for test data. The goal of this experiment is to test the generalizability of a pre-trained classifier when the test data is corrupted in extreme ways. To evaluate, we consider 500 random test examples from the Fashion-MNIST dataset. We corrupt images using the following techniques: \textbf{(a)} \emph{Binary Images}: We transform the images to be binary, using a threshold set at the $50^{th}$ percentile value; \textbf{(b)} \emph{Drop Pixels}: We randomly drop $70\%$ of the pixels in the images and set them to 0, and \textbf{(c)} \emph{Blur + Negative}: We blur with a scale of 25, and invert the images. Examples of images for each of the corruptions are shown in Figure \ref{fig:mnist_tasks}. We compare our approach with PGD, and present results in Table \ref{tab:mnist_robust}. The first striking observation is the performance degradation with corrupted test data ($91.5\%$ to $7\%$), while PGD simply cannot recover images due to the corruptions. On the other hand, \suttl~provides robustness to the classifier, without any additional supervision or data collection.

\subsection{Adversarial Defense}
In this section we discuss how \suttl, by design, can act as a powerful defense against several adversarial attacks. The combination of the surrogate along with the GAN prior can effectively clean adversarial data better than existing methods. In this context, the \suttl~defense can be viewed as a generalization of the recent GAN-based defenses \cite{defenseGAN,ilyas2017robust,santhanam2018defending}, where the surrogate, $\hat{f} = \mathbb{I}$, is assumed to be identity, similar to the PGD baseline described earlier.  The \emph{Cowboy} defense \cite{santhanam2018defending} uses an additional adversarial loss term to the defense-GAN loss \cite{defenseGAN} with the discriminator (We implement it with $\lambda_{adv} = 1\mathrm{e}{-3}$). We consider a wide range of strong attacks to benchmark our defense, and we find that in every single case the \suttl~defense is significantly stronger than existing techniques. While we outperform \textit{Defense-GAN}\cite{defenseGAN}, we retain its advantages -- i.e, the \suttl~defense is a test-time only algorithm that does not require any additional training. We are also entirely unsupervised, and do not need knowledge of the classifier prior to deploying the defense, thus leading to a practical defense strategy.

\noindent \textbf{Experimental details:} We use the pre-trained a CNN classifier described in section \ref{sec:robust} with a test accuracy of $91.50\%$ on the Fashion-MNIST dataset. We design a variety of attacks using the cleverhans toolbox \cite{papernot2016cleverhans}, and test our defense on this classifier for all the following experiments. The performance of \suttl~defense is measured using $500$ randomly chosen test images from the dataset, which are cleaned in batches of size $100$ for efficiency.

\begin{figure}[!htb]
\centering
\subfloat[\small{Universal Perturbations \cite{moosavi2017universal}}]{\includegraphics[trim={0.0cm 0cm 0cm 0cm},clip,width=0.8\linewidth]{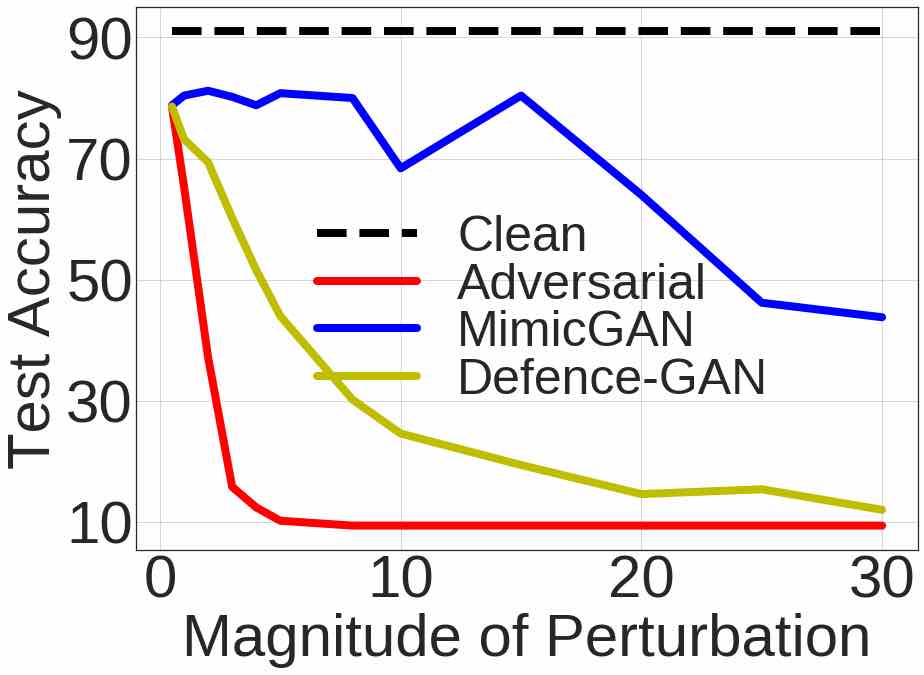}
\label{fig:universal}}

\subfloat[][\small{Popular Adversarial Perturbations}]{
	\centering
	\small
	\begin{tabular}{p{1.5cm}p{1.2cm}p{0.9cm}p{1.3cm}p{1.2cm}}
		\hline

		 \textbf{Attack} &No Defense& Cowboy \cite{santhanam2018defending}&Defense GAN  \cite{defenseGAN}& \suttl\hspace{1pt}\emph{(Ours)}\\ \hline
DeepFool\cite{moosavi2016deepfool}			& 06.20   &  80.20 	& 80.00 	& \textbf{85.40} \\
CWL\cite{CWL2}													& 00.40   &  67.00 	& 66.20 	& \textbf{69.20}\\
BIM\cite{kurakin2016adversarial}				& 05.60 	&  60.00 	& 58.40		& \textbf{68.00} \\
PGDM\cite{madry2018towards}							& 05.40   &  60.80 	& 58.80 	& \textbf{66.40} \\
FGSM\cite{GoodfellowSS14}								& 11.60   &  48.20 	& 46.60		& \textbf{60.00}		\\
Obfuscated\cite{athalye2018obfuscated}	& 23.80 	&  23.80 	& 26.60		& \textbf{33.60} \\ \hline

	\end{tabular}
	\label{fig:adv_table}
	}
	\caption{\small{\textbf{Adversarial Defense:} \suttl~defense shows significantly higher robustness to several strong adversarial attacks. Performance on clean test data is $91.5\%$.}}
    \label{fig:adv}
		\vspace{-15pt}
\end{figure}

\noindent \textbf{Universal perturbations:} \suttl~provides a natural defense against universal perturbations \cite{moosavi2017universal,bbuni}, which are a class of image-agnostic perturbations where an attack is just a single vector which when added to the entire dataset can fool a classifier. To test this defense, we first design a targeted universal perturbation using the Fast Gradient Sign Method (FGSM) \cite{GoodfellowSS14}, by computing the mean adversarial perturbation from $N=15$ test images, i.e. for an adversarially perturbed image $\tilde{X}$, we define the universal perturbation to be $\nu_{u} = \sum_i(X_i-\tilde{X})/N$. We can also increase the magnitude of the attack by scaling it: $\tilde{X}_{u} = X + \alpha \nu_{u}$. Typically, a larger magnitude implies a stronger attack up to a certain point after which it becomes noise and reduces to a trivial attack. In Figure \ref{fig:adv}(a), we observe that when compared to \textit{Defense-GAN}, our defense is significantly more robust.

\vspace{5pt}

\noindent\textbf{Image dependent perturbations:} We test the \suttl~defense against the following image-specific attacks: (a) The Carlini-Wagner L2 attack (CWL) \cite{CWL2}, (b) Fast Gradient Sign Method (FGSM) \cite{GoodfellowSS14}, (c) Projected Gradient Descent Method (PGDM) \cite{madry2018towards}, (d) DeepFool \cite{moosavi2016deepfool}, (e) Basic Iterative Method (BIM) \cite{kurakin2016adversarial} and (f) Obfuscated Gradients \cite{athalye2018obfuscated}. We hypothesize that even though the perturbation on each image is different, the surrogate learns an \emph{average} perturbation when presented with a few adversarial examples. As seen in Table \ref{fig:adv}(b), this turns out to be a strong regularization, resulting in a significantly improved defense compared to baseline approaches such as \textit{Defense-GAN} \cite{defenseGAN} or \textit{Cowboy} \cite{santhanam2018defending}. The obfuscated attack \cite{athalye2018obfuscated} is the strongest attack considered here as it targets the GAN in addition to the classifier. While \suttl~is vulnerable to such an attack, it can afford a stronger defense than plain GAN-based defenses because of the surrogate, as seen in table \ref{fig:adv}(b).

\section{Discussion}
In this paper, we presented \suttl~, an entirely unsupervised system that can recover images from \emph{unknown} corruptions by introducing a corruption-mimicking surrogate in addition to a GAN prior. Such a system moves us a step closer towards a \emph{universal} solution to inverse problems. This can lead to the design of image recovery systems that can be deployed as a pre-processing step for achieving robust classification in the \emph{wild}. \suttl~can recover realistic images over a wide array of extremely ill-posed inversion tasks, while being competitive with supervised approaches. Furthermore, we showed that it improves over recent GAN-based defenses for adversarial attacks, thereby producing state-of-the-art defense against the strongest attacks.
{\small
\bibliographystyle{ieee}
\bibliography{refs}
 \section*{Disclaimer}

 \noindent This document was prepared as an account of work sponsored by an agency of the United States government. Neither the United States government nor Lawrence Livermore National Security, LLC, nor any of their employees makes any warranty, expressed or implied, or assumes any legal liability or responsibility for the accuracy, completeness, or usefulness of any information, apparatus, product, or process disclosed, or represents that its use would not infringe privately owned rights. Reference herein to any specific commercial product, process, or service by trade name, trademark, manufacturer, or otherwise does not necessarily constitute or imply its endorsement, recommendation, or favoring by the United States government or Lawrence Livermore National Security, LLC. The views and opinions of authors expressed herein do not necessarily state or reflect those of the United States government or Lawrence Livermore National Security, LLC, and shall not be used for advertising or product endorsement purposes. }

\newpage
\noindent \textbf{\LARGE{APPENDIX}}
\begin{appendix}
\section*{CelebA Dataset}
\section{Architecture}
In this section we provide implementation and architectural details of the various networks used in this paper. For the CelebA dataset, we use the same generator and discriminator networks specified by DCGAN\cite{radford2015unsupervised} by modifying code available online\footnote{\url{https://github.com/sugyan/tf-dcgan/blob/master/dcgan.py}}. For the surrogate network we use the following network:

\begin{table}[!htb]
\centering
{\small
	\begin{tabular}{ccc}

		\hline
		\textbf{Input}  & \textbf{Operation}\\ \hline
  $(64,64,3)$ & Conv $(5,5,3,16)+ \text{ReLU}$  \\
  $(64,64,16)$ & Conv $(5,5,16,16)+ \text{ReLU}$ \\
  $(64,64,16)$ & Conv $(5,5,16,\text{dims})+ \text{ReLU}$ \\
  $(64,64,\text{dims})$ & Mutiply $(64,64,\text{dims})+ \text{TanH}$  \\

\end{tabular}
\vspace{-5pt}
}
	\caption{\small{CelebA Surrogate}}
\label{tab:celebA_surrogate}
\vspace{-10pt}
\end{table}

\subsection{Corruption Functions}
We use six challenging corruptions to test MimicGAN. In this section we provide implementation details for all of them.

\noindent 1. \textbf{Inpainting:}
\begin{lstlisting}
  import numpy as np
  from np.random import randn

def inpainting(imgs):
    mask = np.ones((64,64,3))
    sig = 0.5
    mask[4:8,12:54,:] = sig*randn(4,42,3)
    mask[12:16,12:54,:] = sig*randn(4,42,3)
    mask[20:24,12:54,:] = sig*randn(4,42,3)
    mask[28:32,12:54,:] = sig*randn(4,42,3)
    mask[36:40,12:54,:] = sig*randn(4,42,3)
    mask[44:48,12:54,:] = sig*randn(4,42,3)
    mask[52:56,12:54,:] = sig*randn(4,42,3)
    mask[60:64,12:54,:] = sig*randn(4,42,3)
    img0 = np.array([x*mask for x in imgs])
    return img0
\end{lstlisting}
\vspace{10pt}

\noindent 2. \textbf{Occlusion:}
\begin{lstlisting}
import numpy as np
from np.random import randn
def occlusion(imgs):
    mask = np.ones((64,64,3))
    mask[24:44,12:54,:] = 0
    mask *= np.transpose(mask,[1,0,2])
    img0 = np.array([x*mask for x in imgs])
    img1 = img0.flatten()
    idx = np.where(img1==0)[0]
    img1[idx] = 0.5*randn(len(idx))
    return img1.reshape(-1,64,64,3)
    \end{lstlisting}
\vspace{10pt}

\noindent 3. \textbf{Pixel Error:}
\begin{lstlisting}
import numpy as np
from np.random import randn

def pixel_error(imgs):
    img0 = np.copy(imgs)
    z = 26
    x = img0[:,:,[z],:]
    noise = randn(imgs.shape[1],z,imgs.shape[3])
    noise = np.tile(noise,[imgs.shape[0],1,1,1])
    _x = np.tile(x,[1,1,z,1])+ 0.25*noise
    img0[:,:,:z,:] = _x
    return img0
    \end{lstlisting}

\vspace{10pt}

\noindent 4. \textbf{Negative:}
\begin{lstlisting}
import numpy as np
from np.random import randn
from cv2 import GaussianBlur as gb

def negative(imgs,scale=25):
    img0 = []

    for i in range(imgs.shape[0]):
        z = gb(imgs[i,:,:],(scale,scale),0)
        img0.append(z)
    return 0-np.array(img0)


    \end{lstlisting}
\vspace{10pt}

\noindent 5. \textbf{Stylize:}
\begin{lstlisting}
import numpy as np
from np.random import randn
import np.percentile as perc
from cv2 import Stylization as style

def stylize(imgs):
    img0 = np.copy(imgs)
    for i in range(imgs.shape[0]):
        src = imgs[i,:,:,:]
        _img = style(127.5*(1+src), sigma_s=10, sigma_r=0.8)
        _img2 = np.array(_img,dtype=np.float32)/127.5 - 1.
        img0[i,:,:,:] = _img2

    img1 = img0.flatten()
    idx = np.where(img1>perc(img0,90))[0]
    img1[idx] =  perc(img0,90) + 0.5*randn(len(idx))
    img_style = img1.reshape(-1,64,64,3)
    return img_style


    \end{lstlisting}

\vspace{10pt}

\noindent 6. \textbf{Gray Blur:}
\begin{lstlisting}
import numpy as np
from cv2 import GaussianBlur as gb
import cv2
def cv_blur(imgsscale=25):
    img0 = []

    for i in range(imgs.shape[0]):
        _x = gb(imgs[i,:,:],(15,15),0)
        y = np.array(127.5*(_x+1),dtype=np.uint8)
        _,_x = cv2.threshold(y,120,255,cv2.THRESH_TRUNC)

        z = np.array(_x,dtype=np.float32)/127.5-1.
        img0.append(z)
      return np.expand_dims(np.array(img0),axis=3)


    \end{lstlisting}

\begin{figure}[!htb]
	\centering
\includegraphics[trim={0.0cm 0cm 0cm 0cm},clip,width=0.95\linewidth]{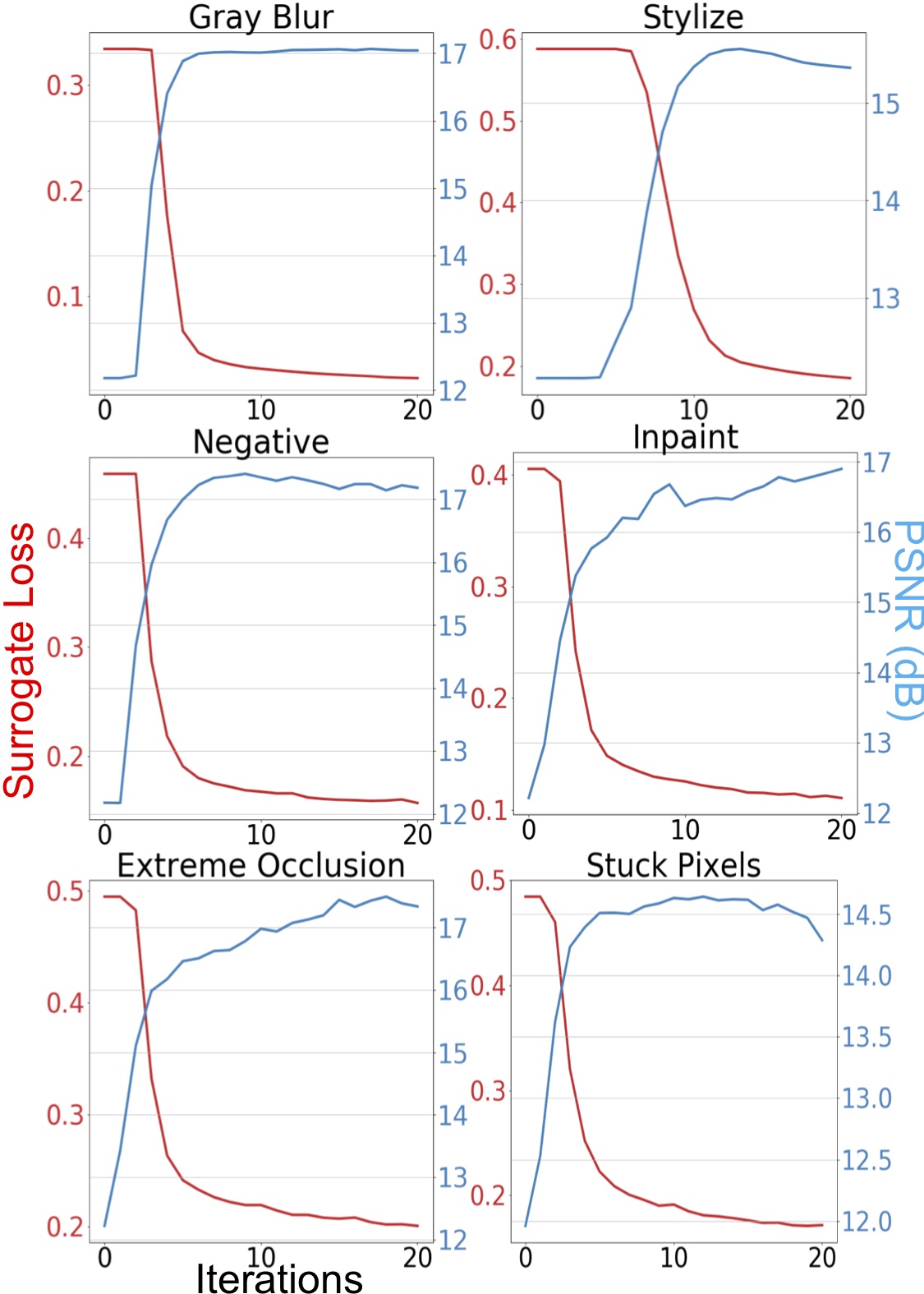}
	\caption{PSNR vs Surrogate loss for all the tasks considered in this work.}
	\label{fig:sup_tasks}
\end{figure}

\section{Additional Results}
\begin{figure*}[!htb]
\centering
\subfloat[\Large{Results for \emph{Gray Blur}.}]{
\includegraphics[trim={0.0cm 0cm 0cm 0cm},clip,width=0.75\linewidth]{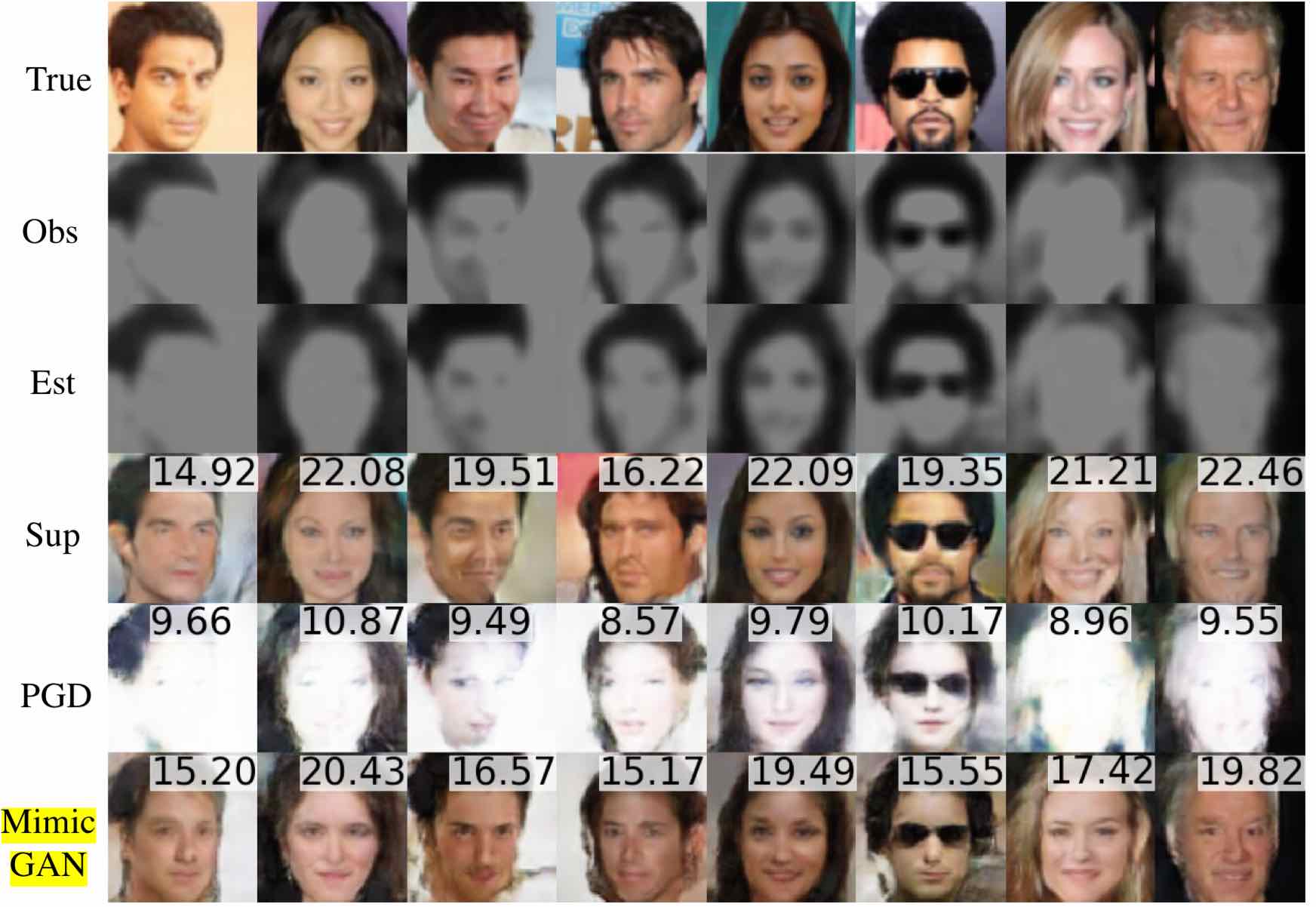}
	\label{fig:sup_gray}}
	\caption{Addtional Results}

\subfloat[\Large{Results for \emph{Stylize}.}]{
\includegraphics[trim={0.0cm 0cm 0cm 0cm},clip,width=0.75\linewidth]{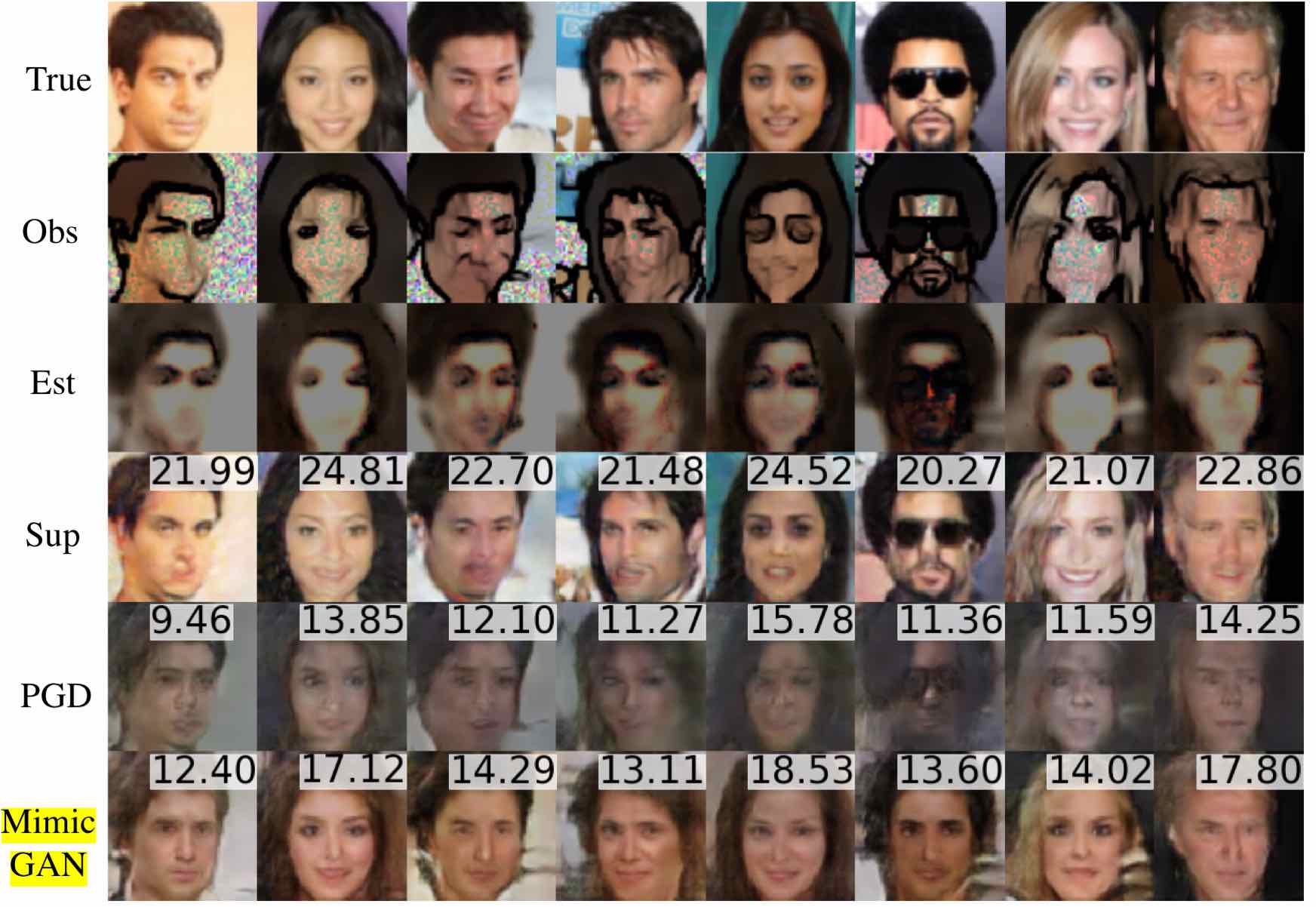}
	\label{fig:sup_stylize}}
\end{figure*}

\begin{figure*}
	\centering
\subfloat[\Large{Results for \emph{Negative}.}]{
\includegraphics[trim={0.0cm 0cm 0cm 0cm},clip,width=0.75\linewidth]{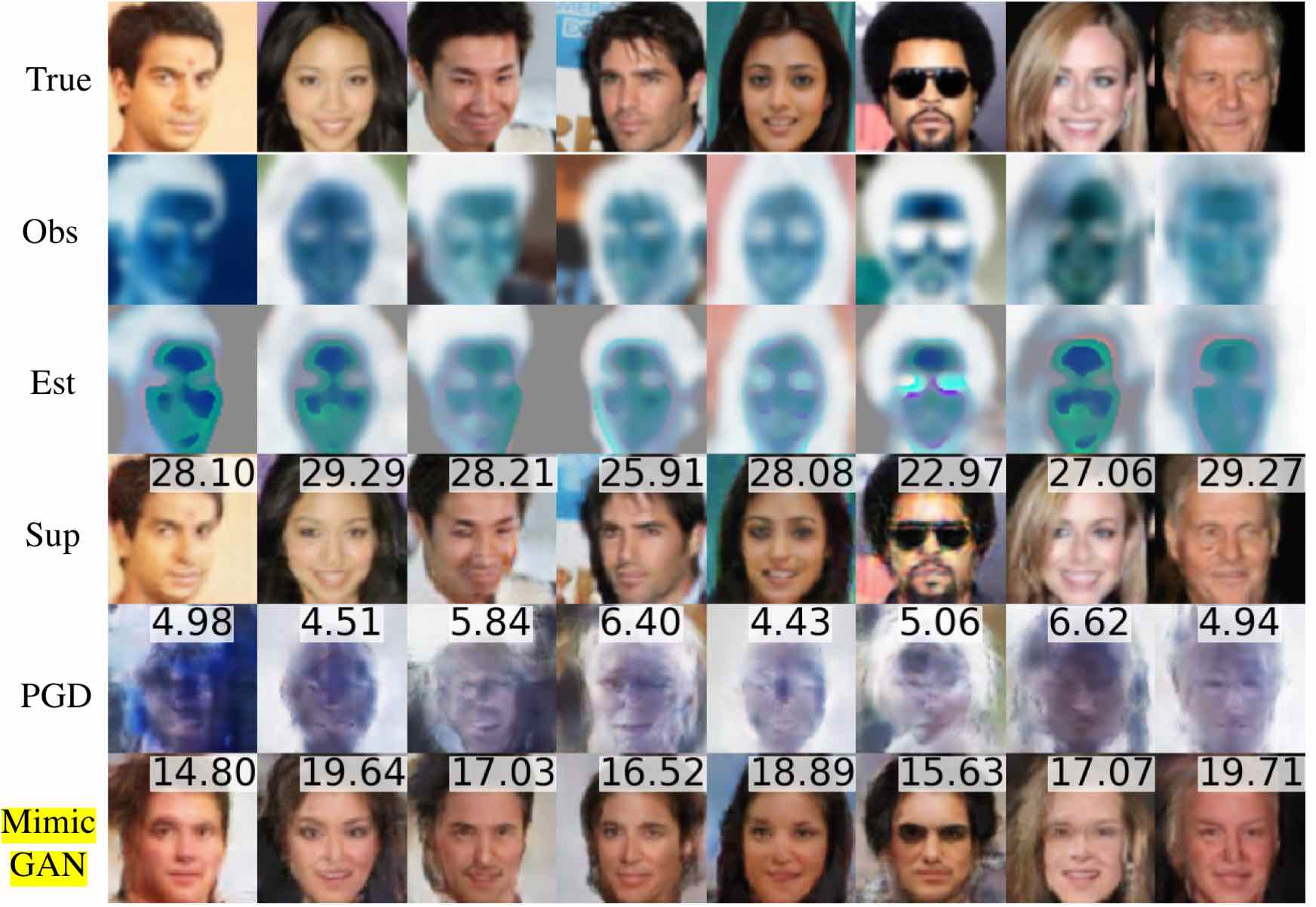}
	\label{fig:sup_negative}}
	\caption{Addtional Results}

\subfloat[\Large{Results for \emph{Pixel Error}.}]{
\includegraphics[trim={0.0cm 0cm 0cm 0cm},clip,width=0.75\linewidth]{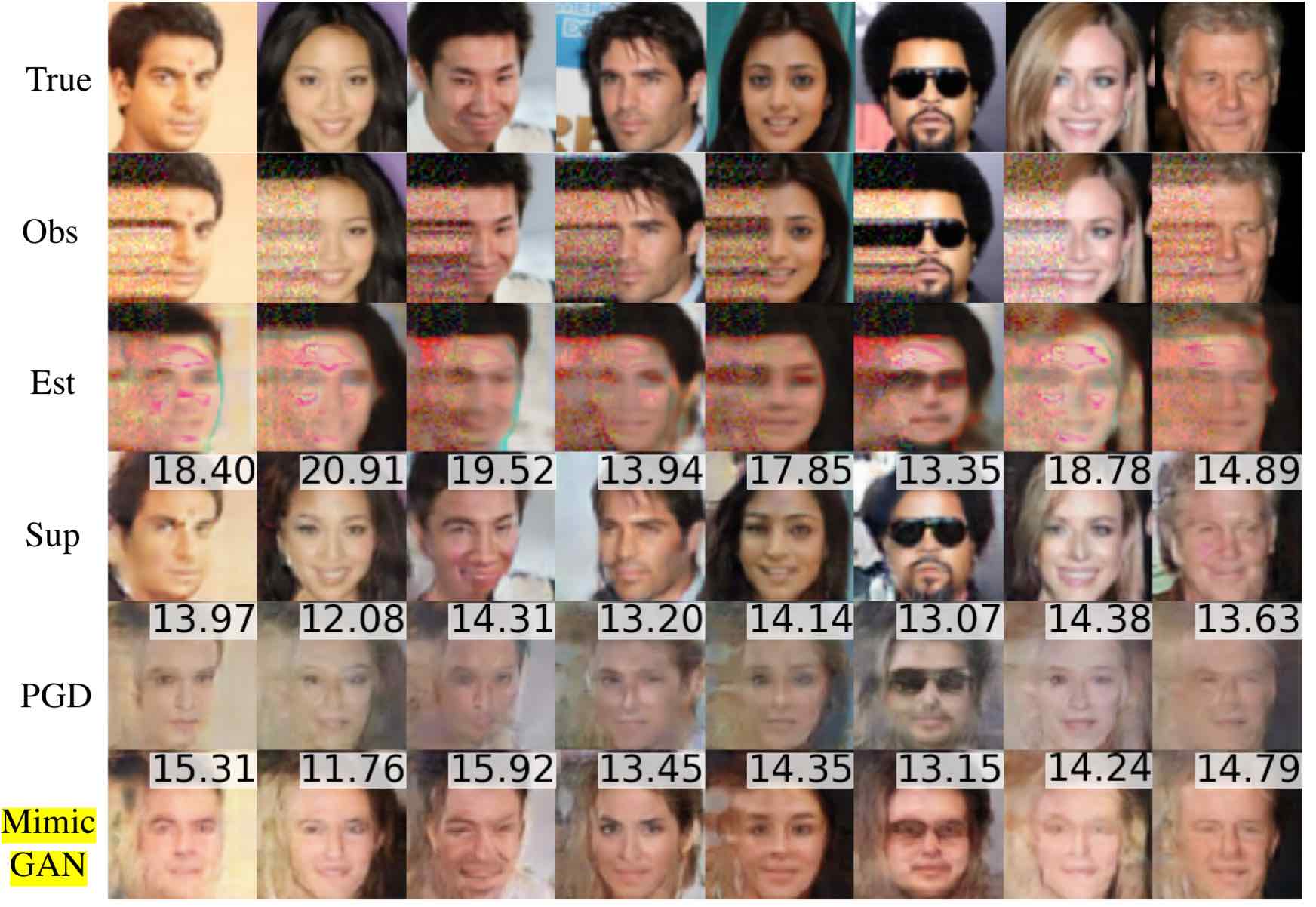}
	\label{fig:sup_pixel}}
\end{figure*}

\begin{figure*}
	\centering
\subfloat[\Large{Results for \emph{Occlusion}.}]{
\includegraphics[trim={0.0cm 0cm 0cm 0cm},clip,width=0.70\linewidth]{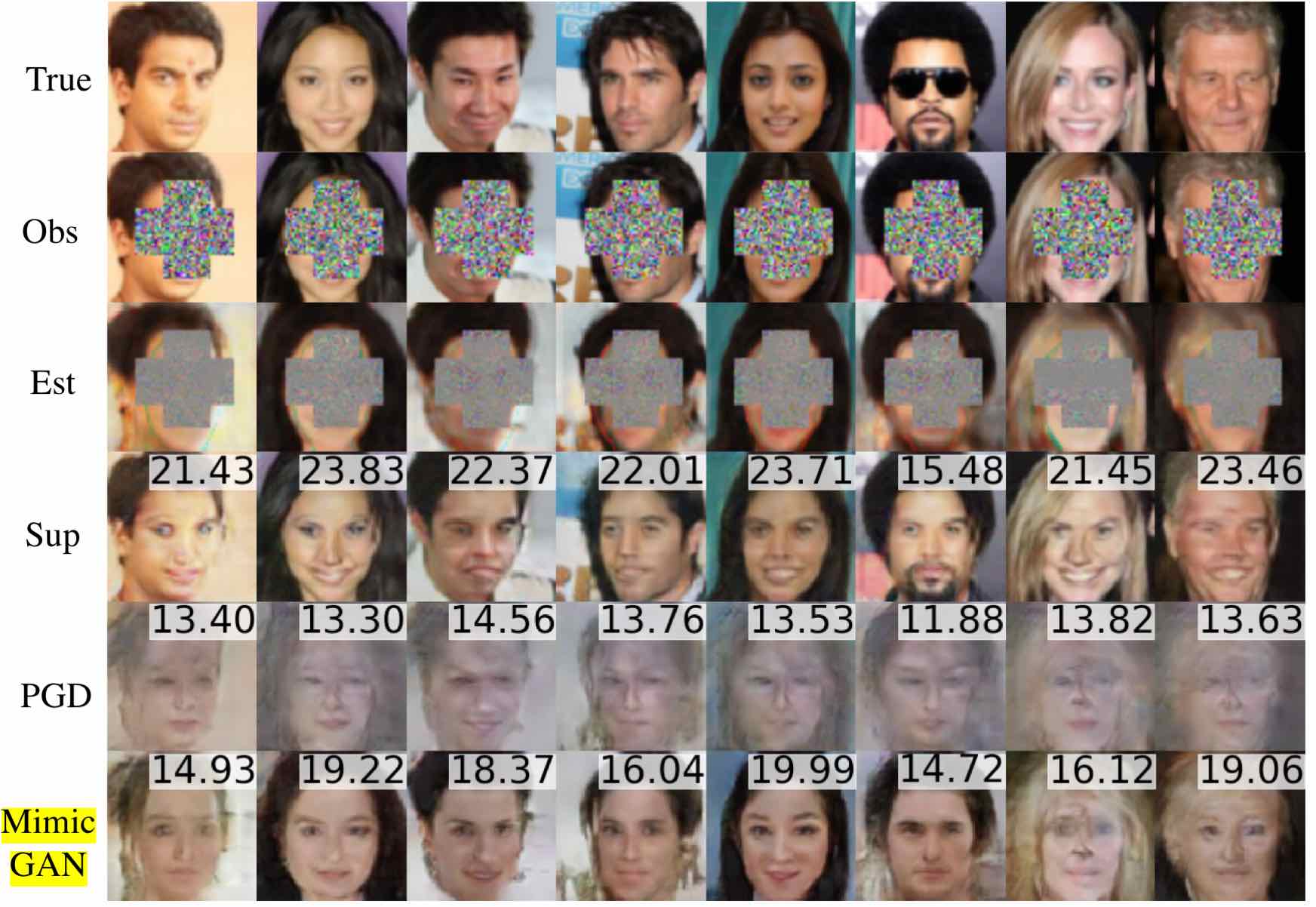}
	\label{fig:sup_occlusion}}
	\caption{Addtional Results}

\subfloat[\Large{Results for \emph{Inpainting}.}]{
\includegraphics[trim={0.0cm 0cm 0cm 0cm},clip,width=0.7\linewidth]{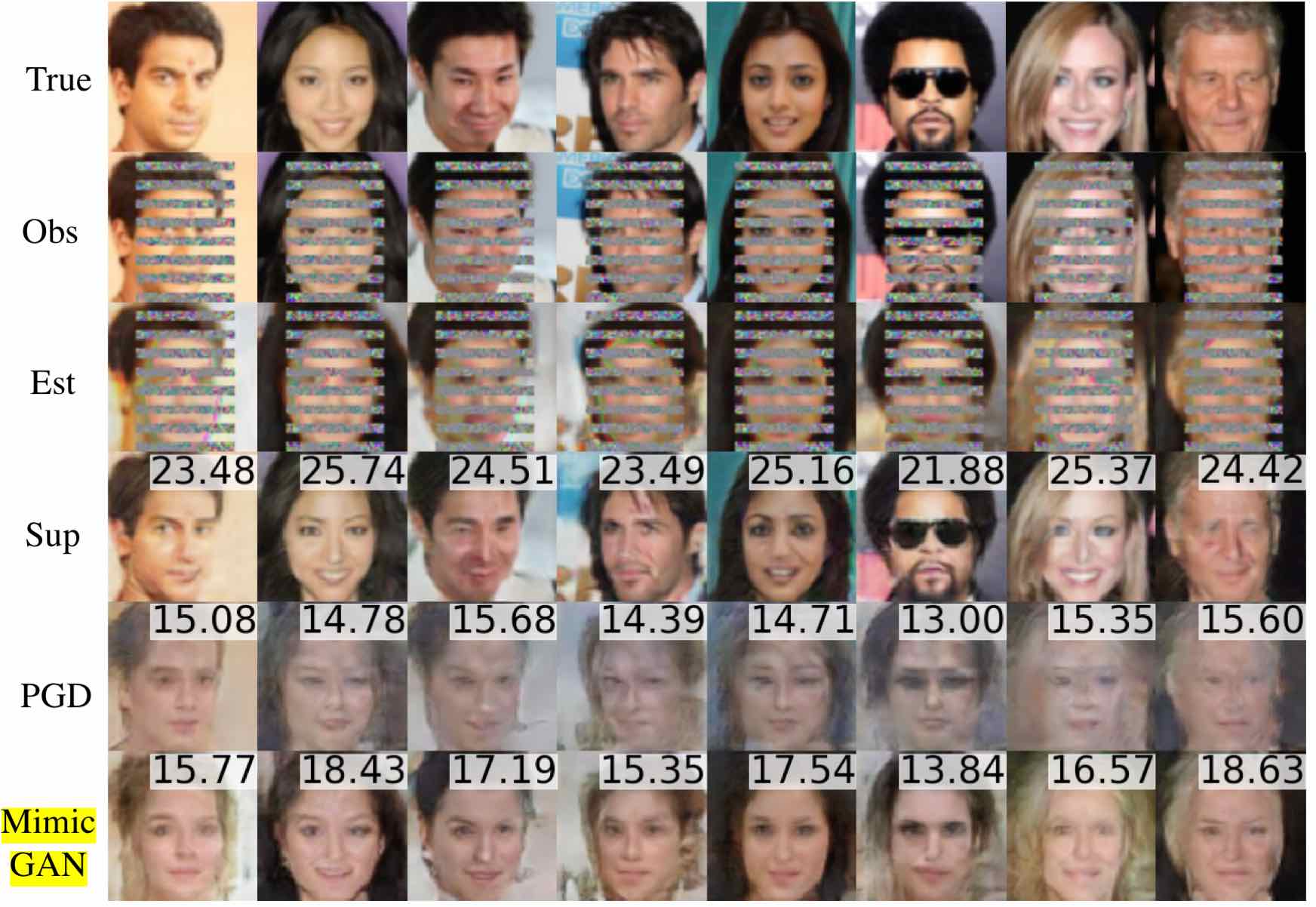}
	\label{fig:sup_inpainting}}
\caption{Addtional Results}
\end{figure*}

Here we show additional results obtained using MimicGAN. In figure (\ref{fig:sup_tasks}), we show how the surrogate improves as the solution estimate is update with each iteration of the alternating optimization loop. In figures (\ref{fig:sup_gray}), (\ref{fig:sup_stylize}), (\ref{fig:sup_negative}), (\ref{fig:sup_pixel}), (\ref{fig:sup_occlusion}), and (\ref{fig:sup_inpainting}) we show additional results for blind image recovery on the six tasks considered here. For definitions of the task please see the experiments section. For all these results, we use the same hyperparameters described in the main text, in table 1.


\section*{Fashion-MNIST Dataset}
\section{Fashion-MNIST}
Here, we outline the different networks used in our Fashion-MNIST experiments. FC is a fully connected layer, BN is the batch normalization operation. In each layer (except masking), we also includes a bias term as well, not shown in the tables.

\begin{table}[!htb]
\centering
{\small
	\begin{tabular}{ccc}

		\hline
		\textbf{Input}  & \textbf{Operation}\\ \hline
  $(1,100)$ & FC $(100,1024)+ \text{ReLU}+\text{BN}$  \\
  $(1,1024)$ & FC $(1024,128*7*7)+ \text{ReLU}+\text{BN}$ \\
  $(1,128*7*7)$ & Reshape \\
  $(7,7,128)$ &Strided Conv $(4,4,64,128)+ \text{ReLU}$\\
  $(14,14,64)$ & Strided Conv $(4,4,1,64)+ \text{TanH}$\\

\end{tabular}
\vspace{-5pt}
}
	\caption{\small{Fashion-MNIST Generator.}}
\label{tab:mnist_generator}
\vspace{-10pt}
\end{table}

\begin{table}[!htb]
\centering
{\small
	\begin{tabular}{ccc}

		\hline
		\textbf{Input}  & \textbf{Operation}\\ \hline
  $(28,28,1)$ & Conv $(4,4,1,64)+  \text{BN}+\text{leaky ReLU}$  \\
  $(14,14,64)$ & Conv $(4,4,1,128)+ \text{BN}+\text{leaky ReLU}$  \\
  $(7,7,128)$ & Reshape \\
  $(1,7*7*128)$ & FC$(128*7*7,1024)$ + leaky ReLU + BN \\
  $(1,1024)$ & FC$(1024,1)$ + Sigmoid \\

\end{tabular}
\vspace{-5pt}
}
	\caption{\small{Fashion-MNIST Discriminator}}
\label{tab:mnist_discriminator}
\vspace{-10pt}
\end{table}

\begin{table}[!htb]
\centering
{\small
	\begin{tabular}{ccc}

		\hline
		\textbf{Input}  & \textbf{Operation}\\ \hline
  $(28,28,1)$ & Conv $(5,5,1,16)+ \text{ReLU}$  \\
  $(28,28,16)$ & Conv $(5,5,16,1)+ \text{ReLU}$  \\
  $(28,28,1)$ & Multiply mask $(28,28,1)$  \\
  $(28,28,1)$ & Add + W*Input  \\

\end{tabular}
\vspace{-5pt}
}
	\caption{\small{Fashion-MNIST Surrogate}}
\label{tab:mnist_discriminator}
\vspace{-10pt}
\end{table}

\end{appendix}

\end{document}